\documentclass[final]{cvpr}

\usepackage{times}
\usepackage{epsfig}
\usepackage{graphicx}
\usepackage{amsmath}
\usepackage{amssymb}
\usepackage{xspace}
\usepackage{multirow}
\usepackage{graphicx}

\newcommand{\SSVRL}{S\textsuperscript{2}VRL\xspace}

\usepackage[pagebackref=true,breaklinks=true,colorlinks,bookmarks=false]{hyperref}

\def\cvprPaperID{2693} % *** Enter the CVPR Paper ID here
\def\confYear{CVPR 2021}
\begin{document}

%%%%%%%%% TITLE
\title{Heterogeneous Contrastive Learning: \\ Encoding Spatial Information for Compact Visual Representations}

\author{
Xinyue Huo\textsuperscript{1,2}\quad Lingxi Xie\textsuperscript{2}\quad Xiaopeng Zhang\textsuperscript{2}\quad Longhui Wei\textsuperscript{1,2}\\
Hao Li\textsuperscript{3}\quad Zijie Yang\textsuperscript{4}\quad Wengang Zhou\textsuperscript{1}\quad Houqiang Li\textsuperscript{1}\quad Qi Tian\textsuperscript{2}\\
\textsuperscript{1}University of Science and Technology of China,\quad\textsuperscript{2}Huawei Inc.\\
\textsuperscript{3}Shanghai Jiao Tong University,\quad\textsuperscript{4}Chinese Academy of Sciences\\
\small\texttt{xinyueh@mail.ustc.edu.cn}\quad\small\texttt{\{198808xc,zxphistory,wlh2568@gmail.com}\\
\small\texttt{lihao0374@sjtu.edu.cn}\quad\small\texttt{yangzijie@ict.ac.cn}\quad\small\texttt{\{zhwg,lihq\}@ustc.edu.cn}\quad\small\texttt{tian.qi1@huawei.com}
}

\maketitle

%%%%%%%%% ABSTRACT
\begin{abstract}
Contrastive learning has achieved great success in self-supervised visual representation learning, but existing approaches mostly ignored spatial information which is often crucial for visual representation. This paper presents \textbf{heterogeneous contrastive learning} (HCL), an effective approach that adds spatial information to the encoding stage to alleviate the learning inconsistency between the contrastive objective and strong data augmentation operations. We demonstrate the effectiveness of HCL by showing that (i) it achieves higher accuracy in instance discrimination and (ii) it surpasses existing pre-training methods in a series of downstream tasks while shrinking the pre-training costs by half. More importantly, we show that our approach achieves higher efficiency in visual representations, and thus delivers a key message to inspire the future research of self-supervised visual representation learning.
\end{abstract}

%%%%%%%%% BODY TEXT
\section{Introduction}
\label{introduction}

Deep learning~\cite{lecun2015deep} opens a new era for visual representation learning, yet most deep learning methods are built upon a large corpus of labeled data. Due to the high costs in data annotation, the community urges new algorithms that can make use of unlabeled data for learning representations, \textit{e.g.}, effectively capturing data distribution in a high-dimensional space. Hence, self-supervised visual representation learning (\SSVRL) has been an important topic~\cite{doersch2015unsupervised,gidaris2018unsupervised,noroozi2016unsupervised} in recent years. In the early age, researchers have shown that deep networks, after pre-trained by some self-supervised tasks (\textit{e.g.}, geometry prediction, colorization, \textit{etc.}), often achieve faster convergence in downstream tasks (\textit{e.g.}, object detection, semantic segmentation, \textit{etc.}) as well as higher recognition accuracy.

Recently, the idea of contrastive learning~\cite{chen2020improved,chen2020simple,he2020momentum,khosla2020supervised,grill2020bootstrap} has largely boosted the performance of \SSVRL. It works by encoding each image into a compact vector and asking the model to discriminate itself (often perturbed by data augmentation) from a large set of other instances. It is believed that to arrive at a high accuracy in instance discrimination, compact semantic information must be extracted, and thus the goal of representation learning is achieved~\cite{dosovitskiy2014discriminative}. To push the limit of semantic learning, researchers have also applied stronger data augmentation techniques and demonstrated the effectiveness in \SSVRL~\cite{anonymous2020contrastive,li2020center}.

\begin{figure}[!t]
\centering
\includegraphics[width=8cm]{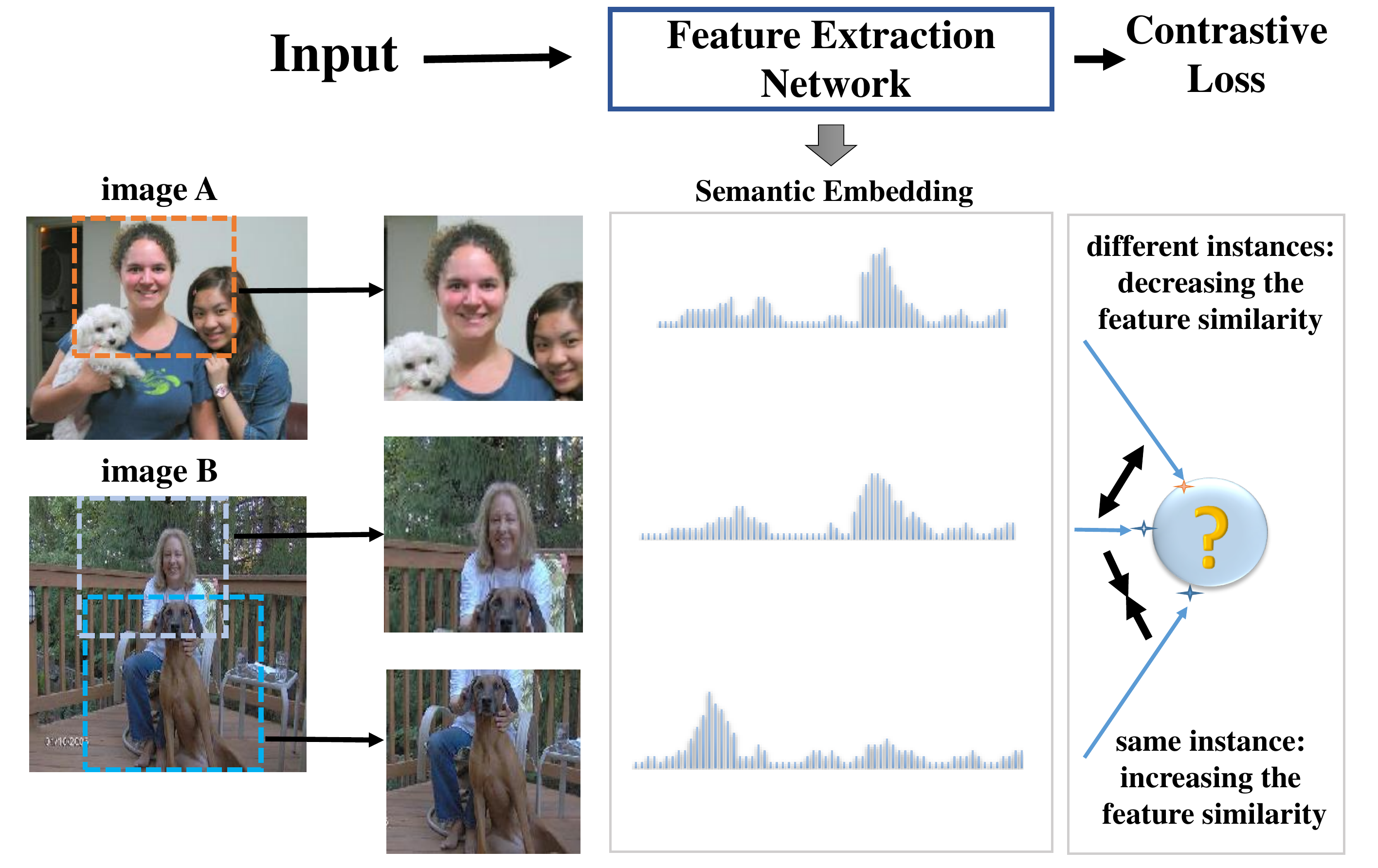}
\caption{Existing contrastive learning methods assumed that all samples from the same image are closely embedded into the feature space. But, under strong data augmentation operations, the samples may have largely different semantics -- sometimes, the sampled patch is closer to other images. This can confuse the learning process in finding compact visual representations.}
\label{fig:motivation}
\end{figure}

But, there is a critical issue that has been mostly neglected. Popular \SSVRL methods such as SimCLR~\cite{chen2020simple} and MoCo~\cite{he2020momentum} do not encode spatial information in the representation vector, yet they require the vector to resist against some data augmentation operations (\textit{e.g.}, image cropping and flipping) that are spatially sensitive. As shown in Figure~\ref{fig:motivation}, this implies that very close representation vectors are assumed to be extracted from an image regardless of which region is cropped and whether it is flipped -- in some extreme cases, the encoded vectors on some regions that have zero overlap on the original image are pulled towards each other by the contrastive objective, which can largely confuse the learning algorithm. In practice, this can downgrade the ability of pre-trained models when they are deployed to some downstream tasks that require spatial understanding, such as object detection and semantic segmentation.

Motivated by the above, we propose an effective approach named \textbf{heterogeneous contrastive learning} (HCL) that appends an individual branch to capture spatial information. In the encoding stage, HCL computes two feature vectors for semantic and spatial information, respectively. The semantic branch is identical to the existing approach that extracts a low-dimensional vector from the last average-pooled layer, followed by a multi-layer perceptron. The spatial branch, for sake of simplicity, follows the feature pyramid network~\cite{lin2017feature}, with the architecture slightly modified to fit the \SSVRL task, to obtain another compact vector (often with a similar length) by integrating multi-stage visual features. These two vectors are then normalized and concatenated before fed into the contrastive loss function. Note that HCL only impacts the contrastive learning stage and the pre-trained model can be deployed to any downstream tasks with no further changes or costs.

We evaluate the effectiveness of HCL in a series of downstream tasks including object detection and semantic segmentation. HCL outperforms all existing \SSVRL approaches in terms of recognition accuracy. In particular, compared to MoCo-v2~\cite{chen2020improved}, HCL uses half of its pre-training costs but surpasses it by APs of $0.8\%$ and $0.7\%$ in MS-COCO object detection and instance segmentation, respectively, and the gain becomes more significant under higher IoU thresholds. We owe such improvement to a more efficient way of learning visual representations. To verify this, we use principal component analysis (PCA) to compress the learned features and perform offline contrastive testing. Under the same dimensionality (\textit{e.g.}, $128$-D), the mixed features (half semantic, half spatial) achieve higher accuracy in instance discrimination. This delivers a key message, possibly more important than the empirical results, that contrastive learning is indeed pursuing high \textbf{efficiency} and low \textbf{information loss} in visual representations, where the former partly reflects in the length of encoded features and the latter is measured by the instance discrimination error. From this viewpoint, HCL explained as heterogeneous features are more efficient visual representations, paving a new path for future research in \SSVRL.

The rest of this paper is organized as follows. Section~\ref{relatedwork} briefly reviews related work, and Section~\ref{approach} describes our approach. After experiments are shown in Section~\ref{experiments}, we draw conclusions and deliver important messages in Section~\ref{conclusios}.

\section{Related Work}
\label{relatedwork}

Deep learning~\cite{lecun2015deep} has achieved great success in a wide range of computer vision problems. The core of deep learning is to build complicated functions named deep neural networks for learning visual representations. In the past years, the optimization of deep neural networks is largely relied on labeled data, for which the community built large-scale image datasets (\textit{e.g.}, ImageNet~\cite{deng2009imagenet}, MS-COCO~\cite{lin2014microsoft}, \textit{etc.}) to maximally cover real-world data distribution. However, there are two major drawbacks of the supervised learning paradigm: (i) the trained models often lack transferability to other domains, and (ii) the ability of exploring unlabeled data is weak. To alleviate these issues, researchers started to investigate unsupervised algorithms for learning visual representations.

This paper focuses on a subarea of unsupervised learning, named self-supervised visual representation learning (\SSVRL). The goal is to capture the data distribution in an unlabeled dataset and (most often) store it in a deep neural network. Though the trained network does not make any semantic prediction, the initialized weights can assist the network training on labeled datasets, often referred to as the downstream tasks. The key of \SSVRL is to find some image nature that corresponds to image semantics to some extent yet does not require any annotations. Typical examples include siamese networks learned from the consistency between the neighboring frames of videos~\cite{fernando2017self,wang2015unsupervised,misra2016shuffle,lee2017unsupervised}, predicting the spatial relationship between image patches~\cite{doersch2015unsupervised} which later evolved into the jigsaw puzzle task~\cite{noroozi2016unsupervised,wei2019iterative,santa2017deeppermnet,noroozi2018boosting} predicting image-level information (\textit{e.g.}, orientation~\cite{pathak2017learning}, rotation~\cite{gidaris2018unsupervised,malisiewicz2009beyond}, counting~\cite{noroozi2017representation}, \textit{etc.}), recovering image contents (\textit{e.g.}, inpainting~\cite{pathak2016context} and colorization~\cite{zhang2016colorful,larsson2017colorization}), \textit{etc}. These efforts achieved accuracy gain on downstream tasks over randomly initialized neural networks, but the performance is far behind the fully-supervised counterpart.

Recently, researchers have opened a new era by noticing that contrastive learning~\cite{he2020momentum} is a promising pretext task for \SSVRL. In contrastive learning, a low-dimensional feature vector is extracted from each image and the goal is to use the feature to identify the image among a gallery that contains a large number of images (\textit{i.e.}, instance discrimination). In practice, the query image is often perturbed by some kind of data augmentation (\textit{e.g.}, image cropping, flipping, \textit{etc.}) and this is partly related to multi-view learning~\cite{wang2015deep,huang2018deepmvs}. To improve the performance, various modifications beyond the raw model have been proposed, including using a multi-layer perceptron to project the features to another space~\cite{chen2020simple}, building a large memory bank to increase the difficulty of instance discrimination~\cite{he2020momentum}, introducing stronger data augmentation operations to challenge the learning algorithm~\cite{tian2020makes,chen2020simple}, \textit{etc}. With these upgrades, the \SSVRL performance in downstream tasks is largely boosted, even surpassing fully-supervised models due to the reduced extent in fitting the source task, \textit{e.g.}, classification. There are also efforts in integrating supervision into contrastive learning~\cite{khosla2020supervised}, showing promising results in learning semantics, but the improvement does not seem consistent. 

\section{Our Approach}
\label{approach}

\subsection{Preliminaries: Contrastive Learning}
\label{approach:preliminaries}

Self-supervised visual representation learning (\SSVRL) starts with an unlabeled image dataset, ${\mathcal{S}}={\left\{\mathbf{x}_n\right\}_{n=1}^N}$, and aims to learn a mapping function, ${\mathbf{y}={\mathbf{f}\!\left(\mathbf{x}\right)}\doteq{\mathbf{f}\!\left(\mathbf{x};\boldsymbol{\theta}\right)}}$, for compact feature representation. In the current era, $\mathbf{f}\!\left(\cdot\right)$ often appears as a deep network and $\boldsymbol{\theta}$ denotes the learnable parameters, \textit{e.g.}, convolutional weights.

Contrastive learning is built upon the assumption that images sampled using different views from the same input image should be represented by similar features. Here, a sampling view indicates a set of operations (\textit{e.g.}, image cropping and flipping) that slightly perturbs the input image. To quantify this request, for each training image, $\mathbf{x}$, two variants are sampled from it, \textit{i.e.}, ${\mathbf{x}_1}={\mathbf{s}\!\left(\mathbf{x},\mathbf{v}_1\right)}$ and ${\mathbf{x}_2}={\mathbf{s}\!\left(\mathbf{x},\mathbf{v}_2\right)}$ where $\mathbf{s}\!\left(\cdot\right)$ is the sampling function and ${\mathbf{v}_1,\mathbf{v}_2}\in{\mathcal{V}}$, the space of sampling views. To force the feature similarity between $\mathbf{x}_1$ and $\mathbf{x}_2$, ${\mathbf{f}\!\left(\mathbf{x}_1\right)}\approx{\mathbf{f}\!\left(\mathbf{x}_2\right)}$, they are put into a gallery, $\mathcal{G}$, that contains a lot other images known as \textit{distractors}, using $\mathbf{x}_1$ as the query, and requiring the network to discriminate $\mathbf{x}_2$ from other instances. This therefore becomes a classification problem and the corresponding loss function can be written as:
\begin{equation}
\label{eqn:contrastive_loss}
{\mathcal{L}\!\left(\mathbf{x}_1,\mathbf{x}_2,\mathcal{G}\right)}={-\mathrm{log}\frac{\mathrm{sim}\!\left(\mathbf{x}_1,\mathbf{x}_2\right)}{\mathrm{sim}\!\left(\mathbf{x}_1,\mathbf{x}_2\right)+\sum_{\mathbf{x}'\in\mathcal{G}}\mathrm{sim}\!\left(\mathbf{x}_1,\mathbf{x}'\right)}},
\end{equation}
where $\mathrm{sim}\!\left(\mathbf{x}_1,\mathbf{x}_2\right)$ is the similarity between $\mathbf{x}_1$ and $\mathbf{x}_2$, often measured by computing the exponential inner-product between the corresponding embedded features and parameterized by the `temperature' hyper-parameter, $T$:
\begin{equation}
\label{eqn:similarity}
{\mathrm{sim}\!\left(\mathbf{x}_1,\mathbf{x}_2\right)}\equiv{\exp\!\left\{\frac{1}{T}\cdot\mathbf{g}\!\left(\mathbf{x}_1\right)^\top\cdot\mathbf{g}\!\left(\mathbf{x}_2\right)\right\}},
\end{equation}
where $\mathbf{g}\!\left(\mathbf{x}\right)$ projects $\mathbf{f}\!\left(\mathbf{x}\right)$ into a low-dimensional space for compacticity. This is often achieved by a multi-layer perceptron upon $\mathbf{f}\!\left(\cdot\right)$.

Recent advances~\cite{he2020momentum,chen2020simple} has revealed several key points to improve the performance of contrastive learning, including using a sufficiently large gallery $\mathcal{G}$ and adding data augmentation to enlarge the gap between $\mathbf{x}_1$ and $\mathbf{x}_2$. Both the modifications can be explained as increasing the difficulty of instance discrimination, so that the algorithm is expected to learn a stronger representation function~\cite{anonymous2020contrastive,li2020center}.

\subsection{Inconsistency between Sampling and Learning}
\label{approach:motivation}

\begin{figure}[!t]
\centering
\includegraphics[width=8cm]{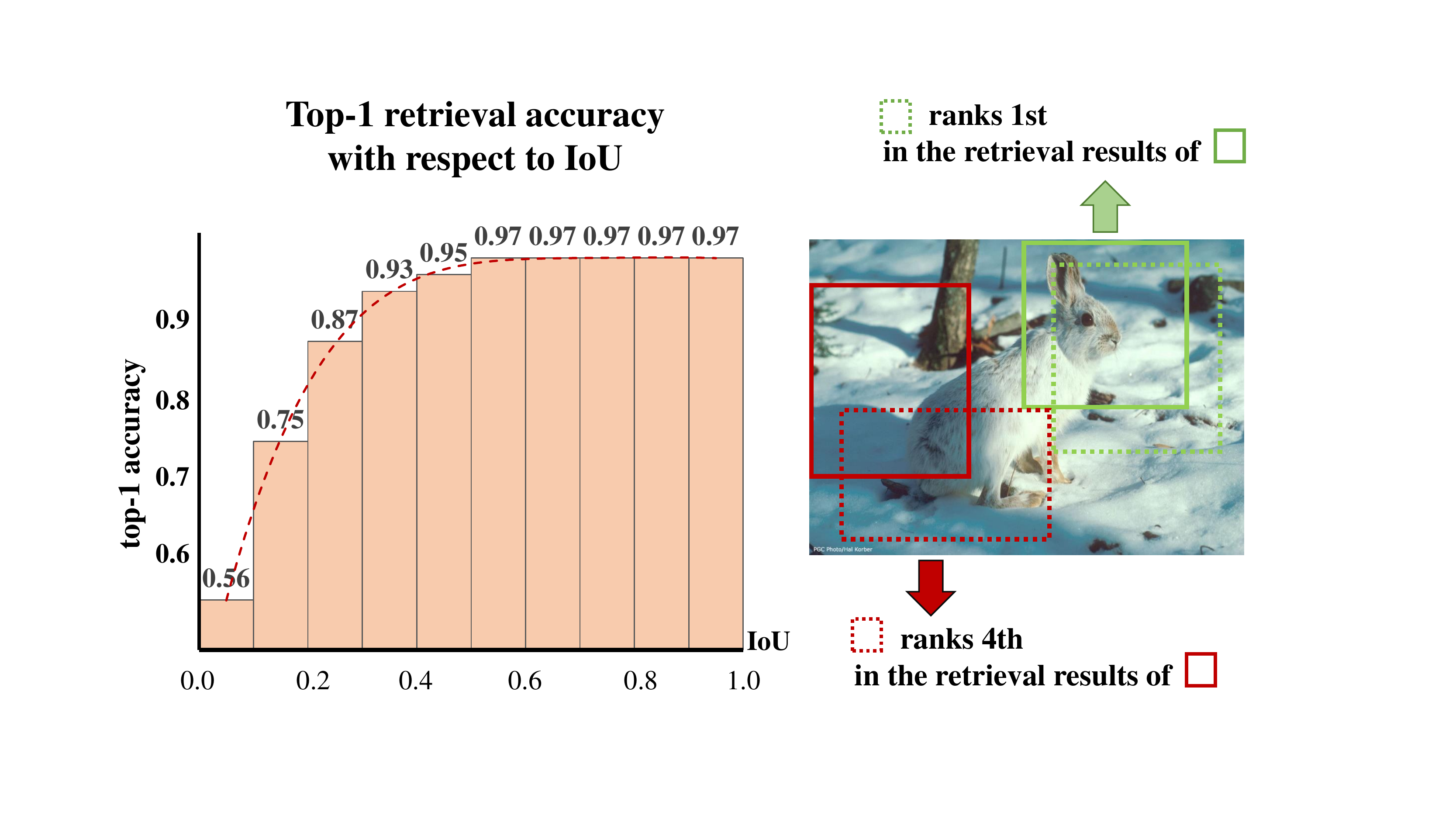}
\caption{\textbf{Left:} the top-1 retrieval accuracy with respect to different correlation coefficients between the two samples. \textbf{Right:} a typical example showing that the retrieval accuracy is closely related to the correlation coefficient.}
\label{fig:correlation}
\end{figure}

The above method has a significant drawback. On one hand, stronger data augmentations are verified effective to enhance the performance of contrastive learning. On the other hand, it becomes problematic to assume that two variants of an image to have close feature representations especially in the scenarios that they are irrelevant due to data augmentation. This is often a difficult task for the learning algorithm.

To verify this statement in statistics, we perform an offline \textit{contrastive testing} to observe how the difference in sampling views affects instance discrimination. Details of contrastive testing are illustrated in Section~\ref{approach:compression}. Briefly, we directly use a ResNet-50 model trained by MoCo-v2~\cite{chen2020improved} for $800$ epochs on ImageNet-1K and traverse through the ImageNet-1K training set, checking whether the top-1 retrieved result is correct. Meanwhile, for each pair of images, $\mathbf{x}_1$ and $\mathbf{x}_2$, we compute the correlation coefficient between the corresponding sampling views, $\mathbf{v}_1$ and $\mathbf{v}_2$, as the intersection-over-union (IoU) ratio of the two sampled rectangles when they are put back to the original image plane. To avoid other factors that impact feature extraction, we only perform image cropping and rescaling during this testing procedure (\textit{i.e.}, horizontal flipping and other color jittering operations are switched off).

Results are shown in Figure~\ref{fig:correlation}. The retrieval accuracy is closely related to the correlation between the sampled views. In other words, using strong data augmentation can deviate the objective of contrastive learning and thus harm the performance of downstream tasks. Intuitively, the difficulty comes from the inconsistency between the data sampling and contrastive learning stages in \textbf{dealing with spatial information}. Specifically, the sampling stage allows different views while the contrastive learning is built upon the encoding algorithm that mostly ignores spatial information.

\subsection{Heterogeneous Contrastive Learning}
\label{approach:solution}

\begin{figure*}[!t]
\centering
\includegraphics[width=16cm]{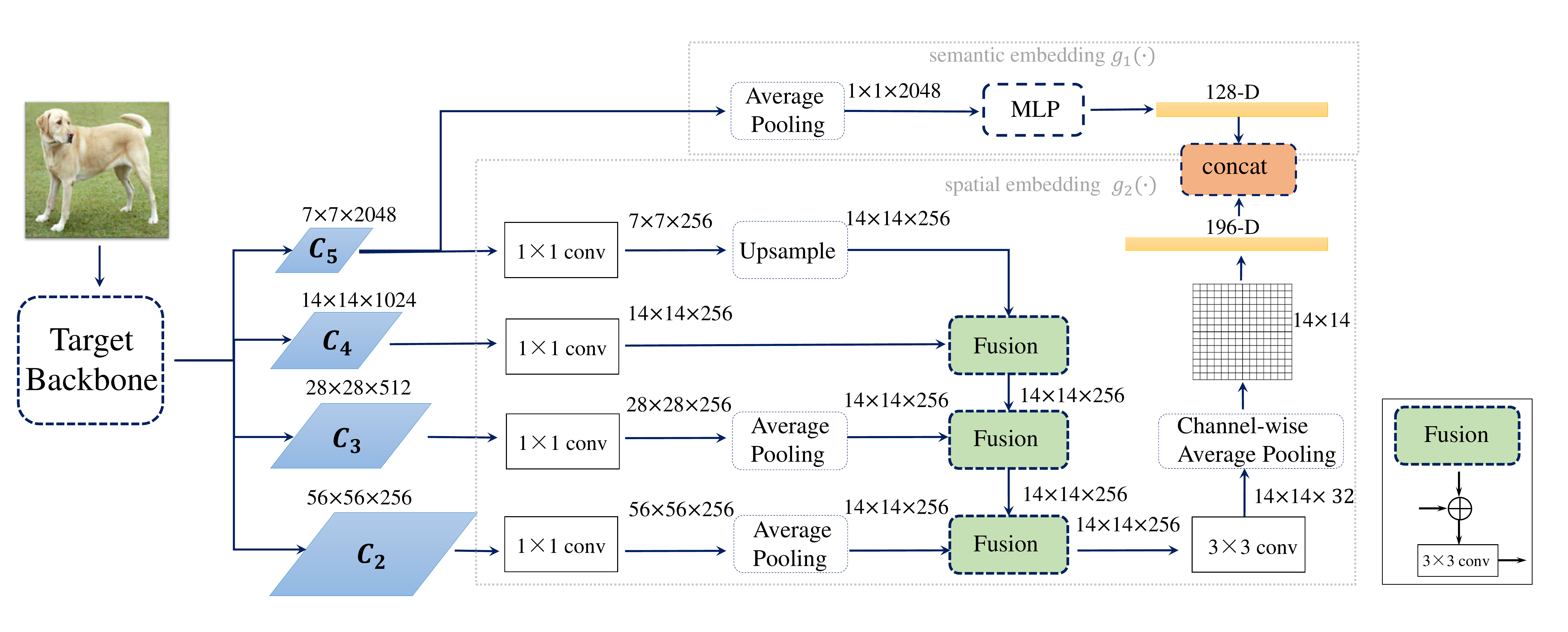}
\caption{The framework of two-branch heterogeneous contrastive learning (HCL). The semantic branch is identical to the head of MoCo-v2~\cite{chen2020improved}, and the spatial branch integrates multi-level features to capture spatial cues. The \textsf{fusion} module is illustrated in the right-hand side. The feature dimensionality is computed based on the ResNet series, yet HCL can be transplanted to other backbones.}
\label{fig:framework}
\end{figure*}

    To overcome the drawback, we explicitly encode spatial information into the embedding vector, $\mathbf{g}\!\left(\mathbf{x}\right)$. The overall framework is shown in Figure~\ref{fig:framework}. Different from the current approach that learns representations from a globally average-pooled feature vector (\textit{i.e.}, with a spatial resolution of $1\times1$), we use an additional vector to capture spatial cues. This is to augment the concept of the embedding function from $\mathbf{g}\!\left(\cdot\right)$ to two individual branches, \textit{i.e.}, $\mathbf{g}_1\!\left(\cdot\right)$ and $\mathbf{g}_2\!\left(\cdot\right)$. For simplicity and convenience of comparison, we assume that $\mathbf{g}_1\!\left(\cdot\right)$ is identical to $\mathbf{g}\!\left(\cdot\right)$ (\textit{e.g.}, in training MoCo-v2 on ResNet-50, it is a two-layer perceptron upon the average-pooled feature from the ${C}_{5}$ layer), and $\mathbf{g}_2\!\left(\cdot\right)$ follows the design of feature pyramid network (FPN)~\cite{lin2017feature} that combines information ${C}_{2}$ through ${C}_{5}$. Similar to computing $\mathbf{g}_1\!\left(\cdot\right)$, there are learnable parameters for extracting $\mathbf{g}_2\!\left(\cdot\right)$ but these parameters will not be used for downstream tasks. Detailed configurations of $\mathbf{g}_1\!\left(\cdot\right)$ and $\mathbf{g}_2\!\left(\cdot\right)$, including the size of each layer, are illustrated in Figure~\ref{fig:framework}. The outputs of $\mathbf{g}_1\!\left(\cdot\right)$ and $\mathbf{g}_2\!\left(\cdot\right)$ are separately normalized and then concatenated for contrastive learning. We name our approach \textbf{heterogeneous contrastive learning} (HCL), since $\mathbf{g}_1\!\left(\cdot\right)$ and $\mathbf{g}_2\!\left(\cdot\right)$ extract different sources of features (\textit{i.e.}, $\mathbf{g}_1\!\left(\cdot\right)$ is mainly designed for semantics while $\mathbf{g}_2\!\left(\cdot\right)$ focuses on spatial information) for visual representation. Throughout the remainder of this paper, we refer to $\mathbf{g}_1\!\left(\cdot\right)$ and $\mathbf{g}_2\!\left(\cdot\right)$ as the semantic branch and spatial branch, respectively.

In our standard implementation, $\mathbf{g}_1\!\left(\cdot\right)$ and $\mathbf{g}_2\!\left(\cdot\right)$ produce $128$-D and $196$-D vectors, respectively, \textit{i.e.}, the final representation vector has $324$ dimensions, about $1.5\times$ longer than the baseline (only using $\mathbf{g}_1\!\left(\cdot\right)$ to extract $128$-D vectors). This brings two impacts. First, each unit (iteration or epoch) of contrastive learning is a little bit ($\sim5.7\%$) slower than the baseline, mainly caused by the additional cost in computing $\mathbf{g}_2\!\left(\cdot\right)$ and calculating Eqn~\eqref{eqn:similarity} in a higher dimensionality. Second, the contrastive learning loss becomes lower (\textit{i.e.}, the instance discrimination accuracy is higher), which is a direct benefit of using auxiliary spatial information for image representation. We will provide an essential explanation on this point in the next subsection. Nevertheless, note that all these changes happen in the contrastive learning procedure. Once it is finished, the pre-trained network can be deployed to any downstream tasks without additional overheads.

Before continuing to the next part, we try to answer an important question: why has HCL alleviated the inconsistency issue of representation learning? We owe this mainly to the contribution of the spatial branch. Assume that $\mathbf{x}_1$ and $\mathbf{x}_2$ are sampled  using largely different views, $\mathbf{v}_1$ and $\mathbf{v}_2$. The conventional contrastive learning algorithms, without the spatial branch, have to force the semantic embedding vectors to be very similar, \textit{i.e.}, pulling $\mathbf{g}_1\!\left(\mathbf{x}_1\right)$ and $\mathbf{g}_1\!\left(\mathbf{x}_2\right)$ together. HCL, equipped with the spatial branch, has an extra opportunity of extracting close spatial embedding vectors. In the scenarios that $\mathbf{v}_1$ and $\mathbf{v}_2$ are weakly correlated (\textit{e.g.}, the sampled images are partly overlapped), it is probable that the convolution and pooling operations in $\mathbf{g}_2\!\left(\cdot\right)$ can learn the correspondence between $\mathbf{x}_1$ and $\mathbf{x}_2$ and thus $\mathbf{g}_2\!\left(\mathbf{x}_1\right)$ and $\mathbf{g}_2\!\left(\mathbf{x}_2\right)$ are sufficiently close. Hence, the contrastive loss, Eqn~\eqref{eqn:contrastive_loss}, is optimized with a weaker constraint on the semantic embedding vectors. The above analysis also explains why the instance discrimination accuracy becomes higher (please refer to the next subsection for examples). However, it is still possible that sometimes, $\mathbf{x}_1$ and $\mathbf{x}_2$ (sampled from the same image) are totally non-overlapped. To solve this challenging case, we need a stronger function for measuring image similarity, \textit{e.g.}, going beyond calculating the inner-product of $\mathbf{g}\!\left(\mathbf{x}_1\right)$ and $\mathbf{g}\!\left(\mathbf{x}_2\right)$. We leave this study for future work.

\subsection{Towards Efficient Visual Representations}
\label{approach:compression}

\begin{figure}[!t]
\centering
\includegraphics[width=7cm]{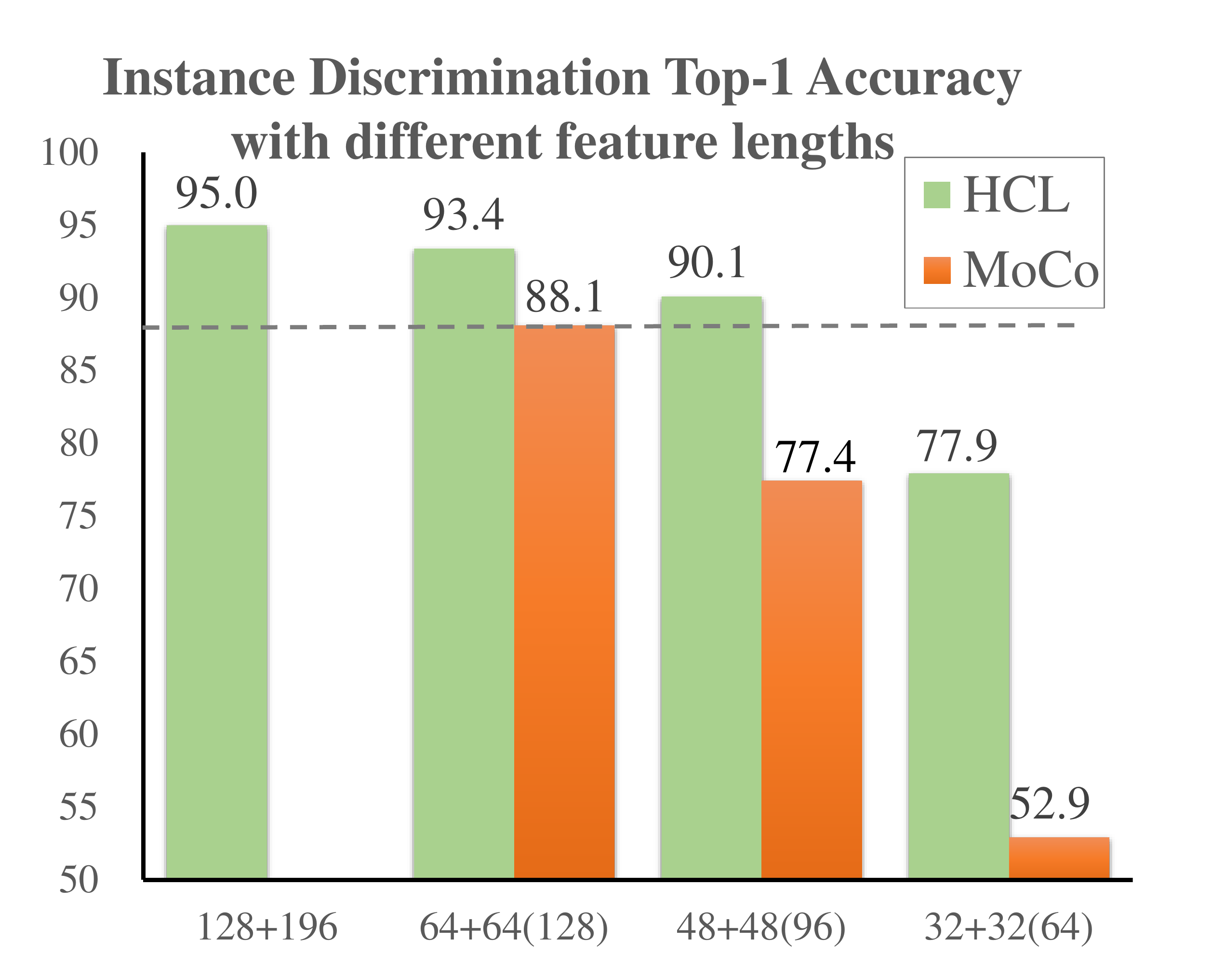}
\caption{The instance discrimination accuracy with respect to different feature lengths, where HCL (mixed semantic and spatial embedding) enjoys significant advantages.}
\label{fig:compression}
\end{figure}

\textbf{This subsection aims to deliver an important message that HCL works better in capturing efficient visual representations.} Intuitively, when each image is represented into a vector of fixed dimensionality, pursuing for a high instance discrimination accuracy is a proper task for \SSVRL. In the contrastive learning framework, the dimensionality of the representation vector serves as a good criterion for the efficiency of representation, and hence we observe the accuracy of instance discrimination with the features of same dimensionality, encoded by different methods.

To ease the evaluation, we design an offline contrastive testing stage where the feature dimensionality can be freely adjusted. We perform a random shuffle on the ImageNet training set and maintain a queue of $65\rm{,}536$ images as the gallery. We perform principal component analysis (PCA) on the extracted features of these reference images and thus project the features to the spaces with the desired dimensionality. The projections for semantic and spatial features are individually computed. The whole training set is used for instance discrimination. Note that the network parameters are fixed, so although the gallery is updated after each iteration, no new learnable information is inserted. In this way, we can report the accuracy of instance discrimination with respect to different augmentation strengths and feature dimensionalities.

We test the standard augmentation strength as used in the training procedure, \textit{i.e.}, a random cropping with image area lying in $\left[0.2,1.0\right]$, random horizontal flipping, and random color jittering. We compare the representations learned from MoCo-v2 (trained on ImageNet-1K for $800$ epochs) and HCL model (trained on ImageNet-1K for $200+200$ epochs -- see Section~\ref{experiments:settings} for details). We test the features under $64$, $96$, and $128$ dimensions. In case of using HCL, each branch contributes half of feature dimensionality. The quantitative results are shown in Figure~\ref{fig:compression}. HCL outperforms MoCo-v2 significantly in instance discrimination accuracy, especially under lower dimensionalities. This delivers the message that semantic and spatial representations are complementary, so combining them achieves higher instance discrimination accuracy. In the next part, we will show that HCL also achieves better performance in a wide range of downstream tasks.

\subsection{Discussions and Relationship to Prior Work}
\label{approach:discussions}

Last but not least, we hope that our investigation inspires researchers in the community to focus on improving the efficiency of visual representations in \SSVRL. Here, we leave some remarks for future research.
\begin{itemize}
\item \textbf{First}, we reveal that the current contrastive learning approach is suffering a conflict between strong data augmentation and accurate instance discrimination. Although introducing spatial information has improved the quality of visual representation, it is still difficult to cover the scenario when the spatial relationship between $\mathbf{x}_1$ and $\mathbf{x}_2$ is weak. To solve this problem under the contrastive learning framework, a better function for similarity computation (\textit{i.e.}, $\mathrm{sim}\!\left(\cdot,\cdot\right)$ in Eqn~\eqref{eqn:contrastive_loss}) should be designed, equipped with the ability to predict that $\mathbf{x}_1$ and $\mathbf{x}_2$ are related even when they do not overlap at all.
\item \textbf{Second}, we point out that instance discrimination accuracy is not a perfect standard for measuring the ability of preserving information. A clear counterexample lies in a learning-free algorithm that extracts a handful of local features (\textit{e.g.}, SIFT~\cite{ng2003sift}) from each image and uses them directly as visual representation. Since these features are robust to scale change, this algorithm may report high instance discrimination accuracy but it rarely helps \SSVRL. We conjecture that a good learning objective should be somewhere between discrimination and generation (\textit{i.e.}, recovering the original image with moderate details), the latter of which has been studied by some researchers~\cite{zhang2019aet,qi2019avt} but the progress seems to fall behind contrastive learning.
%\item \textbf{Third}, we offer an alternative perspective in comparing fully-supervised and self-supervised visual representation learning. We train a ResNet-50 model using full supervision on ImageNet-1K, and compare it to the MoCo-v2 and HCL models in contrastive testing. \textcolor{red}{Although the fully-supervised model achieves higher instance discrimination accuracy \textbf{in the source domain}, the trained model does not transfer well to other domains, \textit{e.g.}, by performing contrastive testing on the MS-COCO dataset, MoCo-v2\textcolor{blue}{($75\%$)} and HCL\textcolor{blue}{$(66\%$, fully supervised $34\%$)} models show advantages in instance discrimination.} This suggests that recent work~\cite{khosla2020supervised} that introduced supervised labels into contrastive learning may benefit from using our diagnostic tool to improve the performance in downstream tasks.
\end{itemize}

\begin{table*}[!t]
\centering
\resizebox{\textwidth}{!}{%
\begin{tabular}{c|c||ccc|ccc|ccc|ccc}
\hline
\multirow{3}{*}{Methods} &\multirow{3}{*}{Data} & \multicolumn{6}{c|}{Mask R-CNN, R50-FPN, detection}                       & \multicolumn{6}{c}{Mask R-CNN, R50-FPN, instance segmentation}                    \\ \cline{3-14} 
                            &  & \multicolumn{3}{c|}{$1\times$ schedule} & \multicolumn{3}{c|}{$2\times$ schedule} & \multicolumn{3}{c|}{$1\times$ schedule} & \multicolumn{3}{c}{$2\times$ schedule} \\ \cline{3-14} 
                 & &  $\mathrm{AP}^{\mathrm{bb}}$ & $\mathrm{AP}_{50}^{\mathrm{bb}}$ & $\mathrm{AP}_{75}^{\mathrm{bb}}$ & $\mathrm{AP}^{\mathrm{bb}}$ & $\mathrm{AP}_{50}^{\mathrm{bb}}$ & $\mathrm{AP}_{75}^{\mathrm{bb}}$ &$\mathrm{AP}^{\mathrm{bb}}$ & $\mathrm{AP}_{50}^{\mathrm{bb}}$ & $\mathrm{AP}_{75}^{\mathrm{bb}}$& $\mathrm{AP}^{\mathrm{bb}}$ & $\mathrm{AP}_{50}^{\mathrm{bb}}$ & $\mathrm{AP}_{75}^{\mathrm{bb}}$\\ \hline\hline
Supervised &IN & $38.9$ & $59.6$   & $42.7$   & $40.6$      & $61.3$ & $44.4$ & $35.4$ & $56.5$ & $38.1$ & $36.8$ & $58.1$ & $39.5$ \\ \hline\hline
MoCo-v1 & IN   & $38.5$ & $58.9$   & $42.0$   & $40.8$      & $61.6$ & $44.7$ & $35.1$ & $55.9$ & $37.7$ & $36.9$ & $58.4$ & $39.7$ \\
MoCo-v2 & IN   & $39.2$ & $59.9$   & $42.7$   & $41.6 $     & $62.1$ & $45.6$ & $35.7$ & $56.8$ & $38.1$ & $37.7$ & $59.3$ & $40.6$ \\
MoCo-v1 &IG  & $38.9$ & $59.4$   & $42.3$   & $41.1$      & $61.8$ & $45.1$ & $35.4$ & $56.5$ & $37.9$ & $37.4$ & $59.1$ & $40.2$ \\ \hline\hline
HCL & IN        & $\mathbf{40.0}$ &   $\mathbf{60.6}$     &  $\mathbf{43.8}$&   $\mathbf{41.8 }$  &   $\mathbf{62.4}$      &    $\mathbf{45.7}$      &  $\mathbf{36.4}$    &  $\mathbf{57.6 }$   &   $\mathbf{39.1}$   &  $\mathbf{37.8} $   &  $\mathbf{59.5}$   &     $\mathbf{40.8}$ \\ \hline
\end{tabular}%
}
\caption{ Object detection and instance segmentation fine-tuned  Mask R-CNN with the R50-FPN backbone on MS-COCO (averaged by 3 runs). IN in tables means model pre-trained on ImageNet-1K~\cite{deng2009imagenet} dataset (1M images) and IG  represents Instagram dataset~\cite{mahajan2018exploring} (1B images).  }%The results of HCL reported in the table are the mean of three runs.
\label{tab:COCO_fpn}
\end{table*}

\begin{table*}[!t]
\centering
\resizebox{\textwidth}{!}{%
\begin{tabular}{c|c||ccc|ccc|ccc|ccc}
\hline
\multirow{3}{*}{Methods} &\multirow{3}{*}{Data} &\multicolumn{6}{c|}{Mask R-CNN, R50-C4, detection}                       & \multicolumn{6}{c}{Mask R-CNN, R50-C4, instance segmentation}                    \\ \cline{3-14} 
                       &  & \multicolumn{3}{c|}{$1\times$ schedule} & \multicolumn{3}{c|}{$2\times$ schedule} & \multicolumn{3}{c|}{$1\times$ schedule} & \multicolumn{3}{c}{$2\times$ schedule} \\ \cline{3-14} 
               & & $\mathrm{AP}^{\mathrm{bb}}$ & $\mathrm{AP}_{50}^{\mathrm{bb}}$ & $\mathrm{AP}_{75}^{\mathrm{bb}}$ & $\mathrm{AP}^{\mathrm{bb}}$ & $\mathrm{AP}_{50}^{\mathrm{bb}}$ & $\mathrm{AP}_{75}^{\mathrm{bb}}$ &$\mathrm{AP}^{\mathrm{bb}}$ & $\mathrm{AP}_{50}^{\mathrm{bb}}$ & $\mathrm{AP}_{75}^{\mathrm{bb}}$& $\mathrm{AP}^{\mathrm{bb}}$ & $\mathrm{AP}_{50}^{\mathrm{bb}}$ & $\mathrm{AP}_{75}^{\mathrm{bb}}$\\ \hline\hline
Supervised & IN & $38.2$ & $58.2$ &$41.2$ &$40.0$&$59.9$ & $43.1$ & $33.3$ &$54.7$ & $35.2$ &$34.7$ & $56.5$  & $36.9$\\ \hline\hline
MoCo-v1 & IN    & $38.5$ & $58.3$& $41.6$ & $40.7$ & $60.5$ & $44.1$ & $33.6$ & $54.8$ & $35.6$ & $35.4$ &$57.3$ &$37.6$\\
MoCo-v2 & IN    & $39.5$ & $59.1$ & $42.7$ &$41.2$ &$61.0$ & $44.8$ &$34.5$ & $55.8$ & $36.7$ &$35.8$ & $57.6$ & $\mathbf{38.3}$\\
MoCo-v1& IG   & $39.1$ &$58.7$  &$42.2$ & $41.1$ & $60.7$ & $44.8$ & $34.1$ & $55.4$ & $36.4$ & $35.6$ & $57.4$ & $38.1$\\ \hline\hline
HCL & IN        & $\mathbf{39.8}$ &   $\mathbf{59.3}$     &  $\mathbf{43.3}$&   $\mathbf{41.4 }$  &   $\mathbf{61.1}$      &    $\mathbf{45.0}$      &  $\mathbf{34.7}$    &  $\mathbf{56.2 }$   &   $\mathbf{37.0}$   & $\mathbf{35.9}$    &  $\mathbf{57.9}$   &     $\mathbf{38.3}$ \\ \hline
\end{tabular}%
}
\caption{Object detection and instance segmentation fine-tuned  Mask R-CNN with the R50-C4 backbone on COCO (averaged by 3 runs). }%The results of HCL reported in the table are the mean of three runs
\label{tab:COCO_C4}
\end{table*}

\section{Experiments}
\label{experiments}

\subsection{Settings and Implementation Details}
\label{experiments:settings}

We use HCL to pre-train the backbone on the ImageNet-1K~\cite{deng2009imagenet} dataset and then fine-tune the pre-trained model in downstream tasks. ImageNet-1K contains around $1.28\mathrm{M}$ training images in $1\rm{,}000$ classes. The validation set containing $50\mathrm{K}$ images is not used. We do not use the semantic labels during the pre-training stage, and merely rely on the instance discrimination task to extract powerful features that are transferrable to the downstream tasks.

Following the setting of MoCo-v2~\cite{chen2020improved}, we choose the ResNet-50~\cite{he2016deep} as the backbone and configure a memory bank with a size of $65\rm{,}536$. There are two stages in training the HCL model. The first stage, known as the warm-up process, is identical to MoCo-v2 and lasts $200$ epochs. After that, the second stage trains the full HCL model for another $200$ epochs. In the second stage, we adjust the initial learning rate to $0.05$ with the cosine annealing schedule and adopt the SGD optimizer with a momentum of $0.9$. The mini-batch size is $256$ in both stages. The two-stage design is to make sure that the network has a basic ability in capturing semantic features -- without it, the training procedure often suffers turbulence. The numbers of epochs in both stages are important hyper-parameters for HCL, which we will diagnose in the later experiments. Under the standard $200$-epoch setting, the two training stages take around $50$ and $53$ hours on eight NVIDIA Tesla-V100 GPUs, respectively.

Note that both stages of pre-training need to be performed only once and can be deployed to a series of downstream tasks just like the models trained with full supervisions. Although HCL introduces a FPN-like architecture to capture the spatial information, the pre-trained weights on these layers are not inherited by the downstream tasks. This is to make a fair comparison to other pre-trained models, in particular, MoCo-v2.

%regular implementation details: backbone, memory bank, epochs, number of GPUs, time cost

%IMPORTANT: how to choose the pre-trained MoCo model, and how many further epochs to perform - to be diagnosed in the ablation part.

\subsection{Performance in Downstream Tasks}
\label{experiments:downstream}

We evaluate our approach for object detection and semantic/instance segmentation, and compare the performance to supervised pre-training and MoCo-v1/v2. The used datasets include MS-COCO~\cite{lin2014microsoft}, LVIS~\cite{gupta2019lvis} , PASCAL VOC~\cite{everingham2010pascal}, and Cityscapes~\cite{cordts2016cityscapes}. For a fair comparison, all settings in the downstream tasks are same as that in MoCo-v1/v2.

\subsubsection{MS-COCO: Detection and Segmentation}

We first evaluate HCL on MS-COCO, a popular benchmark for object detection and instance segmentation. We use the Mask R-CNN~\cite{he2017mask} framework that accomplishes detection and segmentation simultaneously, and equip it with the pre-trained ResNet-50 backbone and either the FPN or C4 heads. All the network layers are fine-tuned on the COCO2017 training set, and we follow the convention to report the average precision (AP) on the validation set under different learning schedules (known as $1\times$ and $2\times$).

The results using FPN and C4 heads are summarized in Table~\ref{tab:COCO_fpn} and Table~\ref{tab:COCO_C4}, respectively. One can observe that HCL surpasses the competitors consistently. The gains are more significant under the $1\times$ setting, where the FPN-equipped HCL outperforms the corresponding MoCo-v2 by $0.8\%$ and $0.7\%$ in terms of detection and segmentation APs, respectively. Besides, HCL also surpasses the supervised pre-training backbone and the MoCo-v1 model using $1$ billion unlabeled images, demonstrating its higher efficiency in learning visual representations.

It is also interesting to see that under the $1\times$ setting, the advantage of MoCo-v2 over supervised training is not as big as HCL enjoys, which delivers the message that the model converges faster when it is initialized by HCL. We conjecture that such benefits imply the nature that object detection and instance segmentation require the model to have stronger abilities in extracting spatial cues. Therefore, by explicitly encoding spatial information during the pre-training stage, HCL is easily adjusted to these downstream tasks, and has a higher upper-bound though the advantages can shrink as the schedule becomes longer.

%Our results are always superior to other self-supervised models and the fully-supervised model with the Softmax loss. We can find that in a shorter learning schedule fine-tuned on the FPN version, MoCo self-supervised model does not have an obvious advantage over the fully-supervised model. Our self-supervised model improve the accuracy to $40.0\%$ in detection task and $36.4\%$ in segmentation task on AP metric. This phenomenon means that the strong object of classification makes the network neglect more detailed representation which is beneficial to the downstream tasks. We also can find that the advantage of HCL model decreases with the increase of training time, this is because the differences between initialized models are dominated by the latter training.

%What should be noted is  that MoCo train the model for 800 epochs which almost need 201  hours with 8 V100 GPUS. We also evaluate the MoCo model trained for 200 epochs on downstream tasks which reports slight lower results in accordance with expectations. HCL only train for 400 epochs in total but achieves higher accuracy, which means our method is more efficient at convergence, more detail about how to set the training strategy is shown in ablation studies part.

\subsubsection{LVIS: Instance Segmentation}

Next, we evaluate HCL on the LVIS-v0.5 dataset for instance segmentation. Different from MS-COCO in which images fall into common categories with abundant labels, LVIS is closer to the real-world recognition scenario that the objects form a long-tail distribution. Therefore, a good practice in LVIS provides more evidences on transferability of the pre-trained backbone. We follow the convention to use the Mask R-CNN model with FPN, and perform the fine-tuning process with a $2\times$ schedule.

The segmentation results on the validation set are shown in Table~\ref{tab:lvis}. HCL produces comparable results to MoCo-v2, and outperforms the supervised backbone once again. These results once again imply that spatial information is complementary to semantic information, and HCL boosts the potential of the pre-trained backbone.

\begin{table}[!t]
\centering
\begin{tabular}{c|c|c||ccc}
\hline
Methods  & Data  & BN     & $\mathrm{AP}^{\mathrm{bb}}$ & $\mathrm{AP}_{50}^{\mathrm{bb}}$ & $\mathrm{AP}_{75}^{\mathrm{bb}}$ \\ \hline \hline
Supervised &IN & frozen &$24.4$ &$37.5$  & $25.8$   \\
Supervised &IN & tuned  &$23.2$ &  $36.0$& $24.4$   \\ \hline
MoCo-v1 &IN & tuned  & $24.1$ & $37.4$  & $25.5$  \\
MoCo-v2 &IN & tuned  & $25.3$ & $38.4$  & $27.0$  \\
MoCo-v1  &IG & tuned  & $24.9$& $38.2$  & $26.4$  \\\hline
HCL&IN  & tuned  &$\mathbf{25.5}$  &  $\mathbf{38.9}$ &  $\mathbf{27.2}$ \\ \hline
\end{tabular}%

\caption{ Long-tailed instance segmentation fine-tuned Mask R-CNN with the R50-FPN  backbone and $2\times$ schedule on LVIS-v05 (averaged by 3 runs).}
\label{tab:lvis}
\end{table}

\subsubsection{PASCAL VOC: Detection and Segmentation}

We then test the tasks of object detection and semantic segmentation on the PASCAL VOC dataset. For the detection task, we use the Detectron2 library~\cite{wu2019detectron2} to fine-tune Faster R-CNN~\cite{ren2015faster} with a ResNet-50 backbone and a C4 head. Follow the settings of MoCo, the fine-tuning lasts for $24\mathrm{K}$ iterations (the $2\times$ schedule) on the the \textit{trainval} set of VOC\textsubscript{07+12}, and the testing is done on the \textit{test} set of VOC\textsubscript{07}. All the training images are rescaled so that the shorter edge length falls in $\left[400,800\right]$, and the testing images are in the original size. For the segmentation task, we fine-tune FCN~\cite{long2015fully} with the ResNet-50 backbone for $50K$ iterations, use the \textit{train} set of VOC\textsubscript{12} for training and the \textit{val} set of VOC\textsubscript{12} for testing.

Results are summarized in Table~\ref{tab:voc}. For the detection task, HCL outperforms both MoCo and the supervised backbone, especially in terms of the $\mathrm{AP}_{75}^{\mathrm{bb}}$ metric where the advatages over the supervised and MoCo counterparts are $6.1\%$ and $0.9\%$, respectively. This phenomenon aligns with that in MS-COCO experiments. Since $\mathrm{AP}_{75}^{\mathrm{bb}}$ has a higher requirement in localization, we confirm that extracting spatial information in the pre-training stage is helpful to localization. For the segmentation task, HCL surpasses MoCo-v2 by a gain of $0.7\%$, largely shrinking the deficit to the supervised baseline. HCL is the best in terms of the overall detection and segmentation performance.

\begin{table}[!t]
\centering
\begin{tabular}{c|c||ccc|c}
\hline
\multirow{2}{*}{Methods} & \multirow{2}{*}{Data} &\multicolumn{3}{c|}{detection} & seg \\ \cline{3-6} 
                        & &$\mathrm{AP}^{\mathrm{bb}}$ & $\mathrm{AP}_{50}^{\mathrm{bb}}$ & $\mathrm{AP}_{75}^{\mathrm{bb}}$   &  mIoU   \\ \hline\hline
Supervised    &IN           & $53.5$  &   $81.4$     &     $58.8$        &   $\mathbf{74.4}$  \\ \hline
MoCo-v1&IN               &  $55.9$ &       $81.5$ & $62.6$            &  $72.5$   \\
MoCo-v2&IN                &   $57.4$ &       $82.5$ &   $64.0$       &  $73.4$   \\
MoCo-v1&IG               &   $57.2$ &       $82.5$ &    $63.7$        &   $73.6$  \\ \hline
HCL&IN    &   $\mathbf{57.9}$     &   $\mathbf{82.7}$     &    $\mathbf{64.9}$    &  $74.1$\\
%HCL&IN* &   $\mathbf{58.0}$     &   $\mathbf{82.9}$     &    $\mathbf{65.2}$    &  $74.3$\\
\hline
\end{tabular}%

\caption{Object detection and semantic segmentation results on PASCAL VOC dataset after fine-tuning  (averaged by 3 runs).}
%* means the HCL model is trained for 800 epochs in the first stage. }
\label{tab:voc}
\end{table}

\begin{table}[!t]
\centering
\begin{tabular}{c|c||cc|c}
\hline
\multirow{2}{*}{Methods} & \multirow{2}{*}{Data} &\multicolumn{2}{c|}{instance seg} & semantic seg \\ \cline{3-5} 
                         &   & $\mathrm{AP}^{\mathrm{bb}}$ &$\mathrm{AP}_{50}^{\mathrm{bb}}$
                         & mIoU\\ \hline\hline
Supervised &IN              &       $32.9$     &      $59.6$  &$74.6$    \\ \hline
MoCo-v1&IN          &   $32.3$    &    $59.3$  & $75.3$               \\
MoCo-v2&IN           &   $33.1$   &     $60.1$     & $75.1$             \\
MoCo-v1&IG               &     $32.9$       &       $60.3$  & $\mathbf{75.5}$   \\ \hline
HCL&IN         & $33.1$          &   $60.2$  &  $75.2$                \\
HCL* & IN     &   $\mathbf{33.6}$   &     $\mathbf{60.8}$    &$\mathbf{75.5}$            \\ \hline
\end{tabular}
\caption{Segmentation results on the Cityscapes dataset (averaged by 3 runs). * indicates that the HCL model is trained for 800 epochs in the first stage.}
\label{tab:city}
\end{table}

\subsubsection{Cityscapes: Instance/Semantic Segmentation}

As the last downstream task, we evaluate the Cityscapes dataset for instance and semantic segmentation in the natural scene images. We use the Mask R-CNN model with FPN as the head and use the $2\times$ schedule. The results of different pre-trained models are summarized in Table~\ref{tab:city}. To alleviate randomness, we repeat the fine-tuning process three times and report the averaged numbers.

The best pre-trained models in this scenario are HCL and the MoCo-v1 model pre-trained on $1$ billion images. This is as expected, since the domain of Cityscapes is largely different from that of ImageNet (auto-driving scenarios, high-resolution), so that the backbone may suffer a domain gap if it is pre-trained \textbf{only} in ImageNet-1K and the spatial information is ignored. MoCo-v1 gains stronger transferability by accessing much more unlabeled image data, yet HCL achieves comparable and even better performance by learning more efficient visual representations, where the latter can save considerable costs in data annotation.

Interestingly, with a longer pre-training stage (the first stage is extended from $200$ to $800$ epochs), the accuracy of HCL continues going up yet MoCo-v2 has arrived at a plateau (with $1\rm{,}000$ epochs, the gain over using $800$ epochs is less than $0.1\%$). We refer the readers to the next section in which we will discuss how the pre-training schedule affects the recognition performance.

\subsection{Diagnostic Studies}
\label{experiments:dignosis}

\subsubsection{Different Training Schedules}
\label{experiments:dignosis:training}

HCL has two pre-training stages, and different training schedules of these stages can affect the recognition performance. Table~\ref{tab:train_epochs} summarizes some comparative results using different numbers of pre-training epochs. From the detection results on the MS-COCO dataset, one can observe that both stages contribute to the recognition accuracy. In particular, (i) skipping the first stage causes a $0.5\%$ drop in both detection and segmentation APs; (ii) undergoing $200$ or $800$ epochs in the first stage does not bring a big difference especially when the second stage is present; and (iii) extending the second stage from $200$ to $400$ epochs leads to consistent accuracy drop.

Integrating the above results yields a big picture of heterogeneous contrastive learning that semantic feature learning serves as a good basis for spatial feature learning, but these two features still seem to conflict because of the different learning goals. Currently, the best choice is to assign $200$ epochs to each stage, so that with a total of $400$ epochs, the pre-trained model surpasses the MoCo-v2 model with $800$ epochs while the computational costs are shrunk by half. Nevertheless, alleviating the conflict and allowing a longer training schedule is important for future research.

\subsubsection{Design of the Spatial Branch}
\label{experiment:module design}

The design of the spatial branch mostly follows the FPN module that integrates four layers of ResNet-50 into the final features. Note that we desire a $14\times14$ map in the end (so that $196$ is comparable to $128$). Unlike the vanilla FPN that integrates all features into a spatial resolution of $56\times56$ (and then down-samples it into $14\times14$), we slightly modify the architecture so that two average-pooling layers are inserted into the third and fourth paths so that the output is directly in $14\times14$. This modification does not affect the recognition accuracy (\textit{e.g.}, on MS-COCO with FPN, the detection and segmentation APs using the vanilla FPN is $39.9\%$ and $36.1\%$, respectively, comparable to our results, $40.0\%$ and $36.4\%$), but thanks to the reduced spatial resolution, our design saves around $10\%$ training cost in the second pre-training stage.

Another important issue is whether to inherit the pre-trained weights in FPN to the downstream tasks, if available. In practice, either option produces similar performance in the downstream tasks, indicating that the initialization of the FPN weights seems less important. For a fair comparison, we switch off this option in the main experiments.

\subsubsection{The Effect of Horizontal Flipping}

In the original contrastive learning baseline, horizontal flipping is a standard operation that forces the backbone to output similar features for an image and its reversed copy. However, when the spatial branch is present, horizontal flipping can cause a mismatch between the spatial features. Not surprisingly, switching off the option of flipping improves the downstream performance (\textit{e.g.}, on MS-COCO with FPN, the detection and segmentation APs with flipping are $39.6\%$ and $36.0\%$, respectively, both are $0.4\%$ lower than that without flipping). The AP of instance segmentation in LVIS-v0.5 also drops by $0.5\%$ with the flipping augmentation switched on. This inspires us that HCL has changed the requirement of data augmentation, and implies the difficulty of integrating semantic and spatial cues into the pre-trained models.

\begin{table}[!t]
\centering
\begin{tabular}{cccc}
\hline
\multicolumn{4}{c}{Mask R-CNN, R50-FPN, $1\times$ schedule} \\ \hline
\multicolumn{1}{c|}{Stage 1} & \multicolumn{1}{c||}{Stage 2} & \multicolumn{1}{c|}{$\mathrm{COCO}_\mathrm{Det}$} & $\mathrm{COCO}_\mathrm{Seg}$  \\ \hline\hline
\multicolumn{1}{c|}{200} & \multicolumn{1}{c||}{0}   & \multicolumn{1}{c|}{38.9}          & 35.4          \\
\multicolumn{1}{c|}{800} & \multicolumn{1}{c||}{0}   & \multicolumn{1}{c|}{39.2}          & 35.7          \\ \hline
\multicolumn{1}{c|}{0}   & \multicolumn{1}{c||}{200} & \multicolumn{1}{c|}{39.5}          & 35.9          \\
\multicolumn{1}{c|}{200} & \multicolumn{1}{c||}{200} & \multicolumn{1}{c|}{40.0}          & \textbf{36.4} \\
\multicolumn{1}{c|}{800} & \multicolumn{1}{c||}{200} & \multicolumn{1}{c|}{\textbf{40.1}} & 36.2          \\
\multicolumn{1}{c|}{200} & \multicolumn{1}{c||}{400} & \multicolumn{1}{c|}{39.7}          & 36.1          \\
\multicolumn{1}{c|}{800} & \multicolumn{1}{c||}{400} & \multicolumn{1}{c|}{39.6}          & 36.0          \\ \hline
\end{tabular}
\caption{The results of pre-trainded models with  different training epochs in the first and second stages on MS-COCO (averaged by 3 runs). The numbers in the top two lines are achieved by MoCo models.}
\label{tab:train_epochs}
\end{table}

\section{Conclusions}
\label{conclusios}

This paper reveals an important problem that spatial information is ignored in the contrastive learning frameworks, so that strong data augmentation including image cropping and flipping can confuse the algorithms, causing a conflict between instance discrimination and representation learning. To alleviate the problem, we propose heterogeneous contrastive learning (HCL) that inserts a branch to capture spatial information and concatenate the semantic and spatial features for contrastive learning. Experiments show that HCL not only achieves better performance in a series of downstream tasks, but also enjoys a higher efficiency in visual representations. We argue that these two factors are closely related, and thus advocate for further study along this direction to improve \SSVRL. In the future, we will also integrate supervised learning into our algorithm~\cite{wei2020can}.

{\small
\bibliographystyle{ieee_fullname}
\bibliography{egbib}
}

\end{document}